\documentclass[10pt,twocolumn,letterpaper]{article}

\usepackage{wacv}
\usepackage{times}
\usepackage{epsfig}
\usepackage{graphicx}
\usepackage{amsmath}
\usepackage{amssymb}
\usepackage{booktabs}
\usepackage[accsupp]{axessibility}

\def\wacvPaperID{1421} 

\wacvfinalcopy 

\ifwacvfinal
\usepackage[breaklinks=true,bookmarks=false]{hyperref}
\else
\usepackage[pagebackref=true,breaklinks=true,colorlinks,bookmarks=false]{hyperref}
\fi
\usepackage{comment}
\begin{document}

\title{Context-empowered Visual Attention Prediction in Pedestrian Scenarios}

\author{Igor Vozniak, Philipp M\"uller, Nils Lipp, Lorena Hell, Ahmed Abouelazm, Christian M\"uller
\and
German Research Center for Artificial Intelligence (DFKI)\\
Saarbr\"ucken, Germany\\
{\tt\small 
\{igor.vozniak,philipp.mueller,nils.lipp,lorena.hell,ahmed.abouelazm,christian.mueller\}@dfki.de}
}

\newcommand{\methodname}{Context-SalNET\xspace}
\newcommand{\methodcubename}{Context attributes\xspace}

\newcommand\philipp[1]{\textcolor{green}{Philipp: #1}}
\newcommand\igor[1]{\textcolor{blue}{Igor: #1}}
\maketitle
\thispagestyle{empty}

\begin{abstract}

Effective and flexible allocation of visual attention is key for pedestrians who have to navigate to a desired goal under different conditions of urgency and safety preferences.
While automatic modelling of pedestrian attention holds great promise to improve simulations of pedestrian behavior, current saliency prediction approaches mostly focus on generic free-viewing scenarios and do not reflect the specific challenges present in pedestrian attention prediction.
In this paper, we present \methodname, a novel encoder-decoder architecture that explicitly addresses three key challenges of visual attention prediction in pedestrians: 
First, \methodname explicitly models the context factors urgency and safety preference in the latent space of the encoder-decoder model.
Second, we propose the exponentially weighted mean squared error loss (ew-MSE) that is able to better cope with the fact that only a small part of the ground truth saliency maps consist of non-zero entries.
Third, we explicitly model epistemic uncertainty to account for the fact that training data for pedestrian attention prediction is limited.
To evaluate \methodname, we recorded the first dataset of pedestrian visual attention in VR that includes explicit variation of the context factors urgency and safety preference.
\methodname achieves clear improvements over state-of-the-art saliency prediction approaches as well as over ablations.
Our novel dataset will be made fully available and can serve as a valuable resource for further research on pedestrian attention prediction.

\end{abstract}

\section{Introduction}

\begin{figure}[t]
  \centering
   \includegraphics[width=1
   \linewidth]{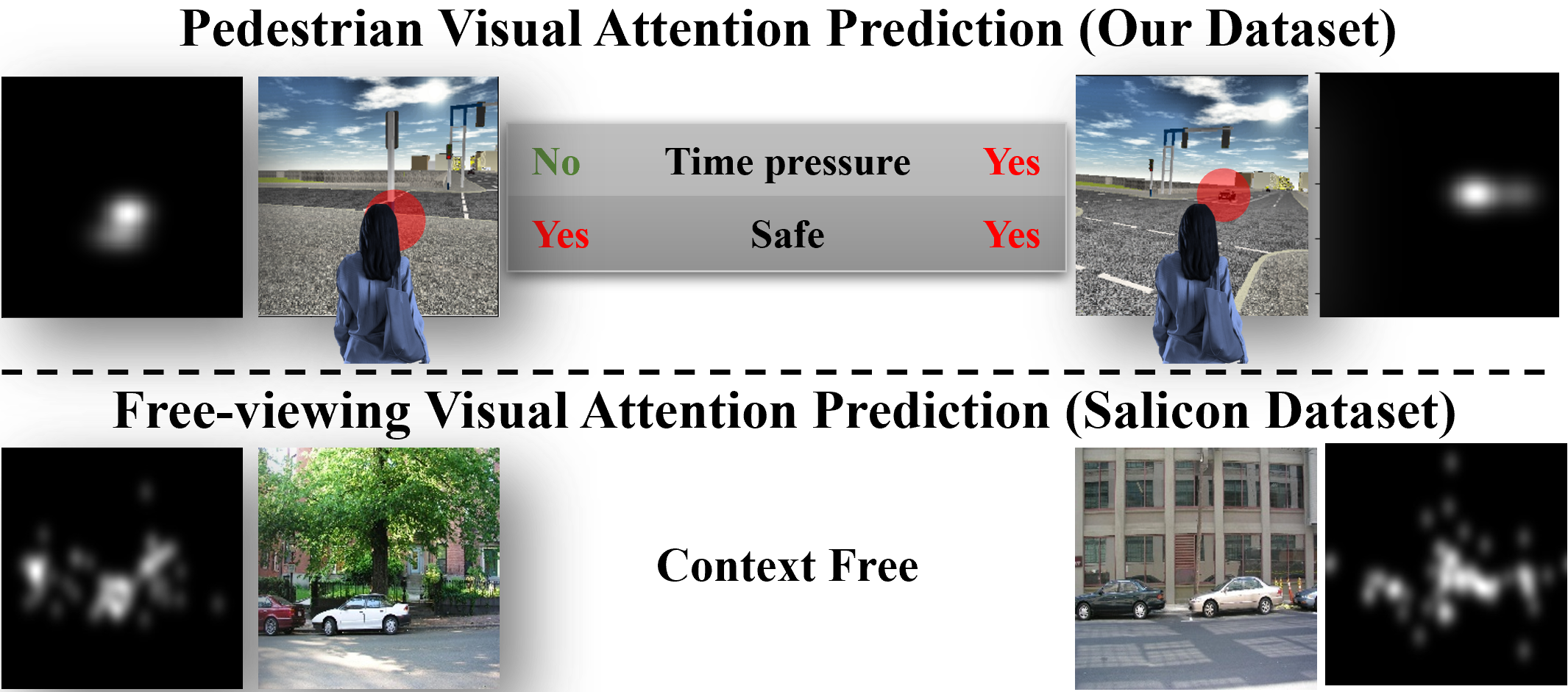}

   \caption{
   In contrast to classical free-viewing visual attention prediction on static images, pedestrian visual attention prediction is highly context dependent. Furthermore, saliency maps generated from pedestrian attention are more sparse compared to free-viewing saliency maps that are aggregated over several subjects viewing the same image.
   }
   \label{fig:teaser}
\end{figure}

The visual behavior of pedestrians in a street crossing situation is influenced by the concrete layout of the street~\cite{de2009pedestrian, leveque2020pedestrians}, but also to a large extent by the existence of the time pressure~\cite{morrongiello2015using, bazilinskyy2021visual}.
Due to its importance to traffic safety, pedestrian attention has been studied extensively in human science~\cite{bazilinskyy2021visual, de2021pedestrians, geruschat2003gaze, tapiro2014visual}.
Automatic prediction of pedestrian attention can open up the possibility to create more realistic training environments both for humans and autonomous agents.
Furthermore, it will help to more accurately model and understand critical traffic scenarios~\cite{homann2018definition}.
Automatic prediction of human attention has received great interest in the computer vision community since more than two decades~\cite{borji2012state,borji2019saliency}.
Significant progress has been made especially on datasets employing a context-agnostic, free viewing paradigm with static images~\cite{itti1998model,kummerer2014deep,droste2020unified}.
These models predict saliency maps that are averages of gaze behavior obtained from several observers for a given static image.
Much fewer works proposed visual attention prediction models in an interactive environment that take into account navigation or search task characteristics~\cite{peters2007beyond,borji2012probabilistic}. 
Until now, no approach for the prediction of pedestrian attention in an interactive environment exists that is able to account for the context factors that are specific to pedestrian behavior (i.e. urgency and safety).
Likewise, to the best of our knowledge, no publicly available dataset to train such a model exists.

We close this gap by proposing the first method and dataset for pedestrian attention prediction in street-crossing scenarios.  Whereas, we do not address the task of salient object detection\footnote{https://paperswithcode.com/task/salient-object-detection}, which is a well-established area.
Our approach consists of an encoder-decoder architecture and addresses three key challenges that distinguish pedestrian attention prediction from the classical scenario of saliency prediction on static images.
First, to capture the context dependence of pedestrian attention, we augment the hidden state of the encoder-decoder with information on the urgency and the safety preference of the pedestrian.
Second, as opposed to the static image scenario, only a few pixels are activated in the saliency maps of visual attention in an interactive environment.
To better cope with this fact, we propose the exponentially weighted Mean Squared Error (\textit{ew-MSE}).
This loss punishes the network less for wrong high-saliency predictions.
Third, neural saliency models are commonly trained on multiple datasets to reduce model uncertainty and achieve the highest performance. 
As only our novel dataset for pedestrian attention prediction is available as of now, we explicitly model the epistemic uncertainty of the model~\cite{kendall2017uncertainties}.

The specific contributions of this work are threefold:
	\textbf{First}, we propose \methodname, the first approach that addresses the task of pedestrian attention prediction. 
	\textbf{Second}, we record the first publicly available dataset pedestrian attention prediction. 
	The dataset consists of diverse street-crossing scenarios recorded in virtual reality and explicitly varies the context factors urgency and safety preference.
	The dataset consists of 528 different scenarios formed based on German In-depth Accident Study (GIDAS) report with a different street layouts and considered in this work context factors. Additionally, the complexity has been extended with layout components like safety-island and multiple lanes in moving directions~\cite{tapiro2018effect, tapiro2020pedestrian}. Thus, the total number of recorded frames is ~35K, which are additionaly labelled with the context information of 11 participants in total. 
	The full dataset will be made publicly available for future research.
	\textbf{Third}, we conduct comprehensive quantitative and qualitative evaluations on this novel dataset, showing the effectiveness of our context modelling approach as well as our proposed ew-MSE loss and the utility of modelling epistemic (statistical) uncertainty.
	In addition, \methodname outperforms a current state-of-the-art saliency prediction approach~\cite{droste2020unified} trained on the same dataset and improves over the current best saliency prediction approach on the MIT/T\"ubingen benchmark~\cite{mit-tuebingen-saliency-benchmark} which was trained on a much larger collection of datasets (no training code available for a direct comparison)~\cite{linardos2021deepgaze}.


\section{Related Work}

Our work is related to the state of the art in human attention prediction, and, more specifically to task-dependent visual attention prediction.

\subsection{State of the Art in Visual Attention Prediction}

Most work on human attention prediction has focused on the task of predicting context-free saliency maps on images~\cite{itti1998model,kummerer2016deepgaze,cornia2018predicting,droste2020unified,tan2019efficientnet}.
The ground truth for this task is a gaze density map averaged over many observers for a given image.
The current state-of-the-art approaches on the influential MIT saliency benchmark~\cite{mit-tuebingen-saliency-benchmark} are DeepGaze IIE~\cite{linardos2021deepgaze} (1st), UniSal~\cite{droste2020unified} (2nd) and SalFBNet~\cite{ding2021salfbnet} (3rd).
DeepGaze IIE improves over its previous version DeepGazeII~\cite{kummerer2016deepgaze} by fusing different backbone networks, thus, the exact training setup is essential to avoid performance bias.
At the time of submission, no open-source implementation of DeepGaze IIE was available that would allow us to train the network on our dataset.
\cite{ding2021salfbnet} proposed SalFBNet which learns a saliency distribution using pseudo-ground-truth, and subsequent fine-tuned on existing datasets.
No implementation was publicly available at the time of submission.
UniSal~\cite{droste2020unified} on the other hand utilizes domain adaptation to train a single model for both image- and video based saliency generation. 
We choose UniSal as a context-free baseline method, as the authors provide an open-source implementation, allowing for training on our dataset. The majority of saliency generation models~\cite{linardos2021deepgaze, droste2020unified, ding2021fbnet} are following similar architectural designs with encoder and decoder. UniSal~\cite{droste2020unified}, for instance, consists of a MobileNet V2~\cite{sandler2018mobilenetv2} encoder, followed by concatenation with learned priors, Bypass-RNN, and a decoder with skip connections, fusion and smoothing layers. The usage of domain-adaptive modules allows for domain-shift between the image and video saliency datasets. 
Note that a large body of work exists on video saliency prediction~\cite{marat2009modelling,wang2019revisiting,zhong2013video,jiang2018deepvs,min2019tased,linardos2019simple,lai2019video}, as well as on egocentric saliency prediction~\cite{tavakoli2019digging,huang2020mutual,yamada2011attention}. 
Recent works in this field commonly extract temporal features like optical flow, recurrences, or 3D convolutions~\cite{min2019tased,linardos2019simple,lai2019video,tavakoli2019digging}.
While these techniques are applicable to our scenario, our focus in this work is to investigate pedestrian attention prediction informed by context attributes, as well as our proposed ew-MSE loss that addresses the challenge of sparse ground truth saliency.
To isolate these aspects and to increase the comparability to the current state of the art in saliency prediction, we choose to leave the integration of temporal features to future work.

\subsection{Task-dependent Visual Attention Prediction}

A large number of works show the importance of the task context in human visual attention allocation~\cite{yarbus1967eye,borji2014defending,land2001ways,hayhoe2005eye,hadnett2019effect}.
For example,~\cite{hadnett2019effect} studied the effects of free-viewing, as well as search- and navigation tasks on visual attention in a virtual environment.They found that navigation, in contrast to free-viewing and search tasks, produces fixations which are more center located. Moreover, in~\cite{land2001ways}, authors studied the relation between eye movements and day-to-day activities like food preparation tasks, indicating nearly all eye movements are made to task-relevant objects. 
It confirms the high effect of the "top-down" component, whereas the bottom-up attributes like color, shape, and size contribute very little to "intrinsic saliency". 
Interestingly, authors in~\cite{brishtel_2022} classified the type of driving (manual vs autonomous) given gaze patterns recorded in a virtual study. All these works scientifically confirm the importance and influence of contextual factors on visual attention. 
%

Despite the importance of context in human attention allocation, only few attention prediction methods explicitly model context. 
An early computational model predicting task-dependent visual attention prediction was introduced in~\cite{peters2007beyond}.
The authors incorporated task-dependent top-down modulation with bottom-up saliency extraction to model participants' attention when playing video games.
Later, \cite{borji2012probabilistic} instructed subjects to navigate in simulated environments (2D and 3D).
Task attributes were modelled by a gist descriptor~\cite{torralba2006contextual,renninger2004scene}, as well as by the subjects' current motor actions.
More recently~\cite{zheng2018task} proposed a task-dependent saliency prediction model for web pages.
Their CNN models task-specific and a task-free aspects of attention in separate branches of the network.
In contrast to previous work which explicitly modelled different kinds of tasks (e.g. navigation versus free viewing~\cite{borji2012probabilistic}), we for the first time explicitly model the qualitative aspects urgency and safety preference within the framework of pedestrian navigation tasks.

\begin{figure*}
  \centering
   \includegraphics[width=0.95\linewidth]{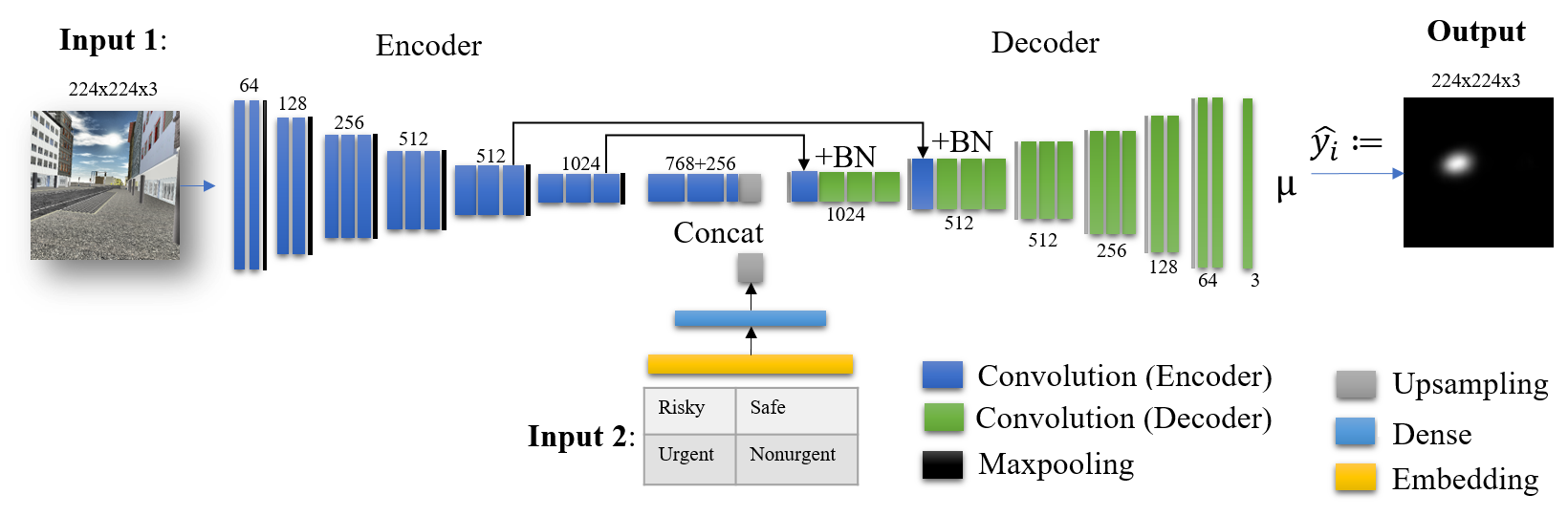}

   \caption{
   CNN encoder-decoder architecture of the generator network. The input consists
of: 1) field-of-view image and; 2) sample of corresponding frame specific context attributes. The objective is to output corresponding attention map. Skip connections are indicated
with arrows between encoder and decoder, where batch normalization (BN) is applied to account for different data distributions.}
   \label{fig:EncDec}
\end{figure*}

\section{Method}
\label{sec:architecture}
The overall architecture of \methodname ~(Figure \ref{fig:EncDec}) consists of an encoder-decoder neural network, which is conditioned on the context attribute information (Input 2, Figure\ref{fig:EncDec}).
To cope with the fact of sparse saliency maps in the interactive pedestrian scenario, we introduce the exponentially weighted MSE (\textit{ew-MSE}) loss.
Furthermore, we model epistemic uncertainty in accordance to~\cite{kendall2017uncertainties} to account for the fact that the available data for pedestrian attention prediction is limited.

\subsection{\methodname Architecture}
Our encoder-decoder architecture is inspired by~\cite{pan2017salgan}, but introduces a novel concatenation layer between encoder and decoder that introduces context information (see \autoref{fig:EncDec}).
%
The encoder consists of blocks of CNN layers.
Each block is followed by a max-pooling layer.
The \textbf{concatenation} bottleneck layer is composed of an embedding layer to encode context information, followed by a fully connected layer with dropout and batch normalization in order to improve the optimization landscape~\cite{santurkar2018does} and to solve for the internal covariate shift. 
The \textbf{decoder} mirrors the encoder except for the addition of upsampling layers to achieve the corresponding resolution. 
In order to maintain fine-grained spatial resolution,
we add skip connections as described in~\cite{he2016deep} between blocks 5 and 6 of the encoder and blocks 1 and 2 of the decoder, respectively. 
In preliminary experiments, these skip connections proved to have a large impact on performance.
%
Except for the Sigmoid output layer, we use ReLU activation functions~\cite{nair2010rectified}. 
The output of the Context-SalNET is a saliency map indicating the attention focus of the pedestrian.
To avoid overfitting and enable probabilistic inference, dropout is applied to blocks 4 and 6 of the encoder and 1-3 blocks of the decoder.

\subsection{Exponentially Weighted MSE Loss}

Compared to classical saliency prediction, where ground truth saliency maps are aggregated over several observers of a static image, ground truth saliency maps in pedestrian attention prediction are much more sparse, only containing few non-zero entries.
To account for this, we modify the mean squared error (MSE) loss that is commonly used in saliency prediction by exponentially weighting it with the magnitude of the prediction.
The resulting exponentially weighted MSE (\textit{ew-MSE}) loss penalises high predictions less, combating the tendency of vanilla MSE to resort to predicting zeros as a result of the sparse ground truth.
Formally,
\vspace{-0.5cm}
\begin{equation}
\centering
\label{eqn:end_dec_loss_ractice}
\begin{split}
 \text{ew-MSE} = \frac{1}{N}\sum\limits_{i=1}^N \exp(-\hat{y}_i)(y_{i}-\hat{y}_{i})^2
\end{split}
\end{equation}
where $\hat{y}$ denotes the model output, $y$ the ground-truth, and $N$ corresponds to the number of output pixels of $\hat{y}$. 

\subsection{Model Uncertainty}
While human attention is influenced by image evidence as well as context factors, the non-deterministic simulation state space including dynamic vehicles, pedestrian, dynamic traffic lights, and obstacles introduces substantial stochastic components. 
Due to the interactive nature of our environment, resulting in different head angles, height, and body orientations, 
every FoV image and corresponding eye-gaze fixation is unique.
In contrast to classical saliency prediction~\cite{itti1998model}, this stochasticity can not be averaged out and the resulting sparseness of data leads to a large model (i.e. epistemic) uncertainty~\cite{hullermeier2021aleatoric}.
To address this challenge, we for the first time propose to model uncertainty in a human attention prediction model. 
We report Epistemic uncertainty inline with \cite{kendall2017uncertainties}, where the dropout variational inference is adopted during the inference phase to approximate the distributions over the network parameters.
Hence, both the training and inference phases are conducted with activated dropout in order to sample from the stochastic posterior, thus, to derive mean and variance over each predicted pixel.
In preliminary experiments, we also evaluated the effects of modelling aleatoric uncertainty, but we observed no performance improvements.

\subsection{Training Details}
We train \methodname according to Equation~\ref{eqn:end_dec_loss_ractice}, where AUC metrics is utilized as an early stop criteria. Given a total of $\sim35K$ pedestrian visual attention images, we set the ratio of 80\% to 20\% for the training-validation data splits,  where the testing is performed on the unseen and subject-specific dataset. 
We utilized Adam \cite{kingma2014adam} optimizer with the loss rate of $10^{-5}$ and the batch size of 96 images across the entire workflow of this work. The input image resolution is set to $224x224x3$, as in line with VGG16 architecture. During leave-one-subject-out cross-validation training, we used clusters with Tesla A100 (40vGB) and Quadro RTAX6000 (48vGb), 2-Core CPU, and 128GB RAM, each. The weights of Encoder (batches of convolution layers 1-5) are initialized from VGG16 \cite{simonyan2014very} for faster convergence and to overcome insufficient gradients.
UniSal~\cite{xu2018personalized} is trained on our dataset in line with the initial training pipeline, allowing for a fair comparison to \methodname. Both UniSal and introduced \methodname rely on backbone networks, ModelNet V2~\cite{sandler2018mobilenetv2} and VGG16~\cite{simonyan2014very} respectively, where both pre-trained on the ImageNet dataset\footnote{https://www.image-net.org/}.

\section{Dataset}
The focus of this research is goal-directed pedestrian behavior in traffic scenarios and the inﬂuence of context attributes, i.e. high-level aspects. Thus, the targets of the research are achievable by utilizing synthetic environments even if a lack in photorealism introduces a domain gap to real images.
However, on a more general note, a domain gap exists in any combination of training vs testing settings~\cite{yang2021st3d} and is beyond the scope of this work.

\subsection{Context Attributes and Scenarios}
We manipulated two context factors in the navigation task that are highly relevant to pedestrian scenarios.
First, we vary the \textit{time pressure} to which participants are exposed.
Second, we instruct participants to perform their task in either a \textit{risky or a safe} way. 
To avoid ambiguity we would like to stress that in our work, we use the notion of \textit{context attributes} (time pressure, riskiness), which differs from the notion of \textit{task} (e.g. free-viewing vs. search vs. navigation) used in some previous works~\cite{hadnett2019effect}. 

\begin{figure*}[]
\centering
  \includegraphics[width=0.95\textwidth]{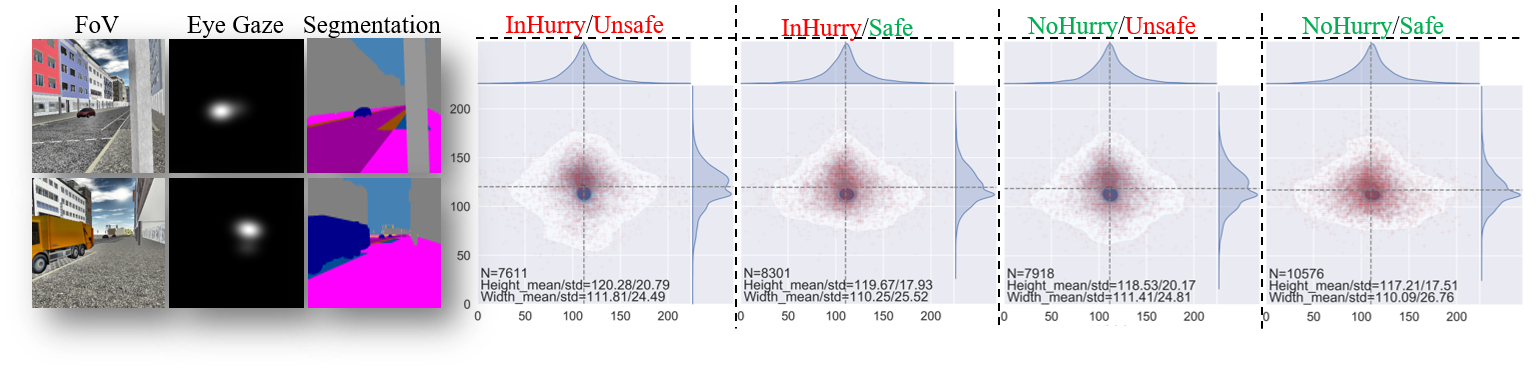}
  \vspace{-0.5cm}
  \caption{Samples of the recorded dataset and additionally extracted information. Left: samples of recorded field-of-view (FoV) images, corresponding eye gaze information, and segmentation maps (inline with CityScapes color schema with extensions caused by fine-grained scene-related details). Right: accumulative distribution of fixation points across all subjects with a context-based split. The $N$ indicates the number of samples of a specific context type, where mean and std values are self-explanatory.
  }
  \label{fig:data_samples}
\end{figure*}

\begin{figure}[]
\centering
  \includegraphics[width=0.48\textwidth]{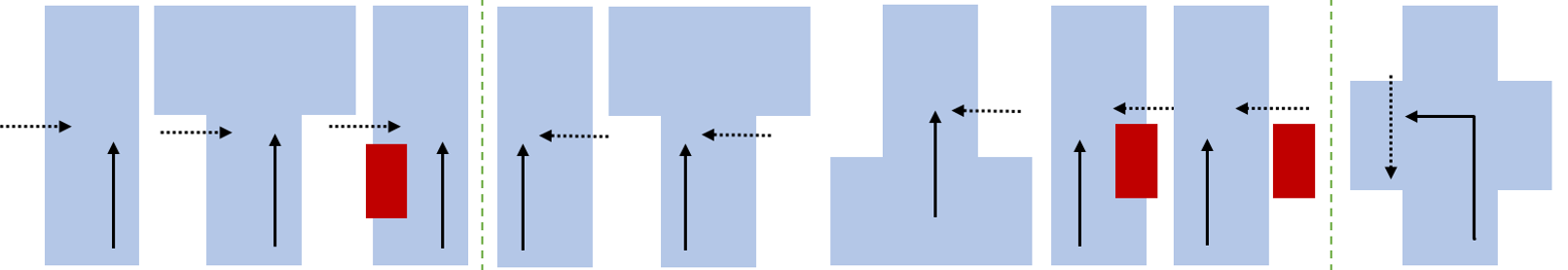}
  \caption{Traffic scenario layouts based on German In-Depth Accident Study (GIDAS). The solid vector stands for the approximated moving direction of the vehicle(s), where the dashed vector indicates an approaching direction of the subject. The red rectangle stands for an obstacle on the way. 
  }
  \label{fig:gidas}
\end{figure}

To record a realistic dataset with a high relevance to challenging real-life traffic situations, we base our scenarios on the German in-Depth Accident Study\footnote{GIDAS - \url{https://www.gidas.org/start.html}} (GIDAS) which identified nine classes of critical street-crossing scenarios (see \autoref{fig:gidas}) that specify street layouts and traffic participants (pedestrian, vehicle, and potential obstacles).
 We added three more additional scenarios to cover additional urban scene complexities like safety island between two opposite direction lanes, multiple lanes in each driving direction, and crossings involving consecutive traffic lights. This helps to additionally increase the variation in participants' visual behavior and the number of opportunities to realise safe or unsafe street crossing behavior.
To further increase the realism, we embedded these scenarios into a virtual reconstruction (digital-twin) of a real city with accurate street layouts including traffic lights, pedestrian street crossings, bicycle lanes, parking spots, as well as reconstructions of the actual buildings.



\subsection{Recording Setup}

To simulate the traffic scenarios we chose the open-source simulation software OpenDS\footnote{OpenDS - \url{https://opends.dfki.de}}, whereas we considered other simulators like Carla\footnote{Carla - \url{https://carla.org/}}, LGSVL\footnote{LGSVL - \url{https://www.svlsimulator.com/}}, and GTA5\footnote{GTA5-\url{https://www.gta5-mods.com/scripts/driving-mode-selection}}, however by the time of conducting the recording session it was missing some important features, e.g., support of pedestrian-centric VR goggles with eye-gaze recording, digital-twin setup and workflow control.  
Moreover, OpenDS has the key advantage that it will allow other researchers to replay the recorded pedestrian trajectories that we plan to publish since the raw as well as post-processed dataset will be released.
This will increase both the reproducibility of our research and the value of the dataset to investigate novel research questions.
Figure~\ref{fig:data_samples} shows samples of recorded images (top row), namely RGB frames with corresponding post-processed saliency and segmentation maps respectively. Besides, the corresponding depth maps are also recorded and to be released. Thus, might assist in future empirical studies. The bottom row in Figure~\ref{fig:data_samples} illustrates distributions of aggregated across all subjects fixation points based on context factors. Moreover, provided visualizations of context based eye-gaze distributions is inline with the empirical studies, where subjects tend to look further away with vertical $mean=120,28$ and $td=20,79$ in case of "InHurry/Unsafe" setup (Figure~\ref{fig:data_samples}, bottom left) to look for more potential hazards like approaching vehicles. Thus, the perception of higher risk leads to more cautions behaviour and more detailed assessment of the traffic before crossing the street. Such factors are i.e., the absence of traffic signals and zebra crosswalks, lower time-to-collision, faster cars, wider streets with several lanes~\cite{rasouli2017understanding}. Whereas, in the case of "NoHurry/Safe" setup (Figure~\ref{fig:data_samples} , bottom right), we observe smallest vertical $std=17,51$ with $mean=117,21$ and highest horizontal $std=26,76$. 
The aim of our study is to model the impact of context attributes on human visual attention as opposed to low-level modelling of fine-grained image features, hence, the rendering capabilities of the gaming engine is not central for our research.

To achieve a maximum degree of realism and immersion in the simulation, we made use of virtual reality goggles. 
We employed the HTC Vive Eye\footnote{HTC Vive Eye - https://www.vive.com/} featuring an integrated eye tracker.
Furthermore, we used two Base Stations 2.0 and collected user input via a Xbox One controller. The camera rotation and translation coordinates are taken directly from VR goggles. Thus,  pitch, jaw, and roll angles as well as actions like jumping or squatting are supported in our setup, which makes it a well-suited benchmark test due to the underlined complexity. 
To balance resolution with simulation performance, 
we choose a sampling rate of 3 frames-per-second to record the subject's current field of view, her current attention, as well as the corresponding semantic segmentation map.


\begin{figure*}[]
\centering
  \includegraphics[width=0.95\textwidth]{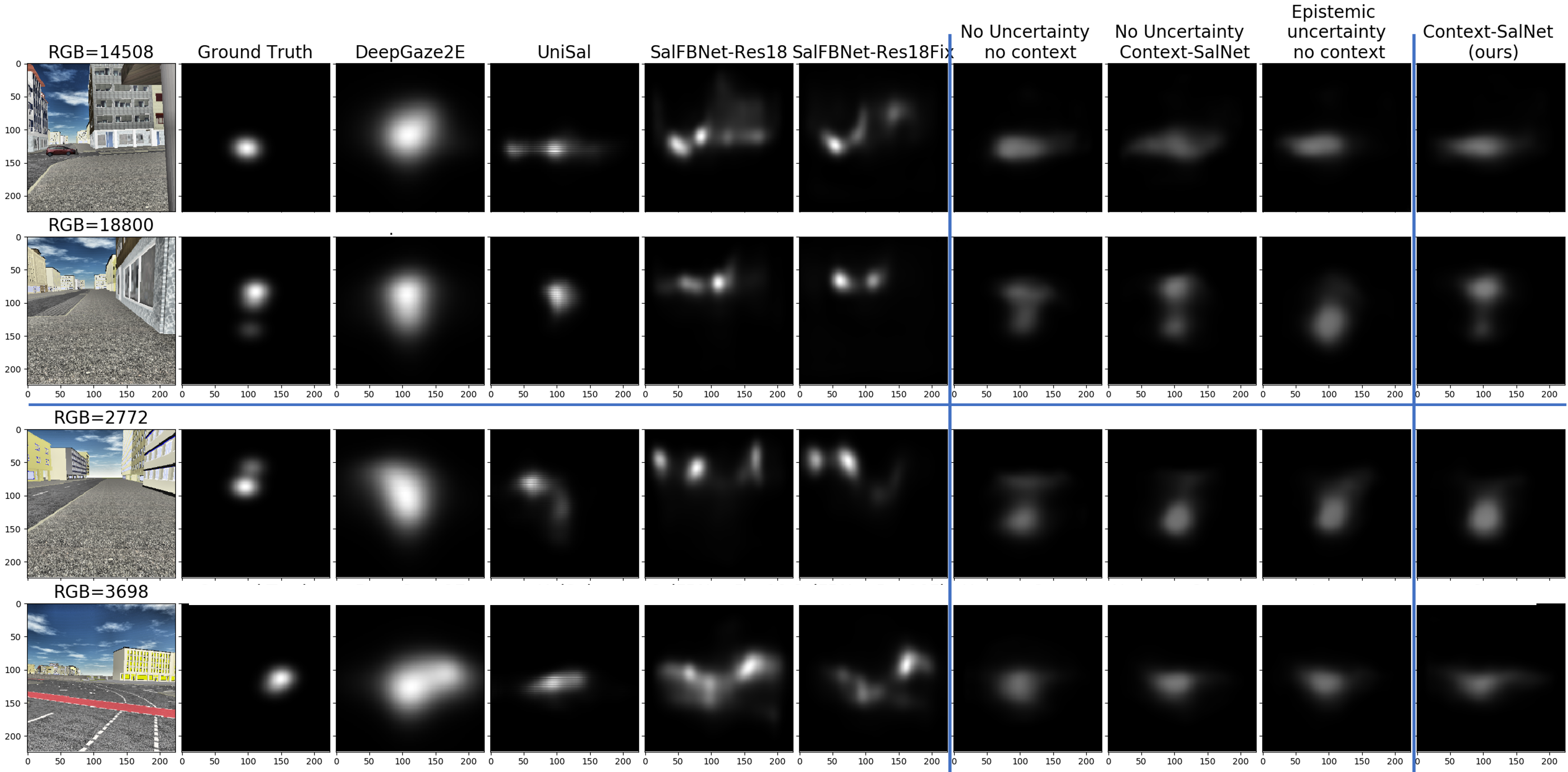}
  \caption{Qualitative analysis of randomly selected best with $(AUC>0.99\%)$ and
worse $(AUC<0.70\%)$ samples. The rows 1-2 stand for the best samples, while rows 3-4 correspond to the worse visual predictions. Provided renderings serve two purposes: 1) qualitative baseline evaluations with columns 2-5, and 9; 2) qualitative ablation evaluations with columns 6-9. Column 1 stands for the input FoV image with a unique RGB sequence ID.} 
  \label{fig:best_worse}
\end{figure*}

\subsection{Procedure}
We recruited 15 participants out of which four withdrew due to feelings of motion sickness.
%
Prior to the study, all participants gave informed consent for participation and for inclusion of their pseudonymized data in the dataset. For each participant, the eye tracker was calibrated at the start of the recording session.
Subsequently, participants spent 5 minutes in the simulation to familiarize themselves with the controls. 
%
%
Participants were presented with four blocks of all 12 traffic scenarios each.
Each block realized one combination of time pressure (yes/no) and riskiness (high/low).
Each of the 12 trials in a block started by visually indicating the target location for 5 seconds.
Data recording started after these 5 seconds have passed.
In each trial, participants are able to move \textit{forward, backward, left and right} and head movements were mapped to camera movements along \textit{pitch, yaw, and roll} angles.
Hence, we collected a unique dataset with 528 scenarios, resulting in $\sim35K$ of unique FoV images and corresponding segmentation, saliency, depth maps as well as xml files to store simulated related information e.g., position, speed, orientation of the body, and head. 

\vspace{-0.25cm}

\section{Experiments}
\label{sec:experiments}
\subsection{Pre-Processing}
Saliency ground truth information consists of fixation sequences, namely recorded $X$ and $Y$ coordinates projected to the image plane. 
Following previous work that made use of fixation maps in pedestrian navigation scenarios~\cite{Vozniak1}, we aggregate the gaze locations obtained from the last three frames to create a representation of participants' current focus of attention.
To arrive at continuous ground truth attention maps, we follow the saliency map computation in~\cite{wang2018revisiting} with \textit{degree of visual angle} set to $dva=9.3$. We discount the previous attention points in intensity to allow the neural network to account for previous information, but to also overfitting to the additional auxiliary information.
On the images recorded from the simulator, we applied Contrast Limited Adaptive Histogram Equalization (AE)~\cite{zuiderveld1994contrast} to obtain even color distributions across images, which improves invariance to unique attributes of the scene, e.g. uniquely colored buildings.
%
%
For optimal alignment with community standards, the labelling scheme of our segmentation maps matches the CityScapes~\cite{cordts2016cityscapes} labelling convention except where we had to introduce new classes that are missing in CityScapes (e.g. bicycle lanes, parking slots).

\subsection{Quantitative Evaluation}
\label{subsec:eval_1}

\begin{table*}[htb]

\centering
\begin{tabular}{llllllll}
\toprule
Method & AUC-J $\uparrow$ & s-AUC $\uparrow$ & AUC-B $\uparrow$ & NSS $\uparrow$ & SIM $\uparrow$ & CC $\uparrow$ & KLDiv $\downarrow$\\ 
\midrule

       DeepGaze2E~\cite{linardos2021deepgaze}            &   0.9526
        &    0.6313
          &    \textbf{0.7842}
        &  2.7158
   &    0.3726
        &  0.5146
  &   0.1326
    \\
    SalFBNet-R18~\cite{ding2021salfbnet}                     &      0.9050
     &   0.5418
          &      0.5818
      &  1.7376
   &       0.2761
     &  0.3225
  &   0.2393
  \\
SalFBNet-R18Fix~\cite{ding2021salfbnet}                 &      0.9014
     &   0.5340
          &      0.5591
      &  1.6121
   &       0.2605
     &  0.2902
  &   0.2646
    \\
\midrule

Center Bias                       &      0.8360
     &    0.5101
          &      0.5381
      &  1.0940
   &       0.2231
     &  0.2130
  &   \textbf{0.1322}
    \\
   UniSal~\cite{droste2020unified}            &   0.9388 
        &    0.5631
          &    0.5961
        &  2.7097
   &    0.3978
        &  0.4537
  &   0.3755
    \\
\midrule


\textbf{\methodname (ours) } &  \textbf{0.9605}
          &    \textbf{0.6654} 
         &    0.7723
       & \textbf{3.3048}
    &    \textbf{0.4646}
        & \textbf{0.5843}
  &   0.1690
    \\

\bottomrule

\end{tabular}
\newline
\caption{Leave-one-subject-out baseline evaluation results using different evaluation metrics.
Arrows indicate whether higher ($\uparrow$) or lower ($\downarrow$) is better.
DeepGaze2E/SalFBNet is shown separately as it is composed of several backbone networks and trained on different training data than the other approaches.
Bold numbers indicate the best results.
}
\label{tab:results1}
\end{table*}

\begin{table*}[htb]
\centering
\begin{tabular}{llllllll}
\toprule
Method & AUC-J $\uparrow$ & s-AUC $\uparrow$ & AUC-B $\uparrow$ & NSS $\uparrow$ & SIM $\uparrow$ & CC $\uparrow$ & KLDiv $\downarrow$\\ 
\midrule

\textit{No uncertainty} &  
          &     
         &     
       & 
    &    
        & 
  &   
    \\
    
\ \ vanilla $MSE$, no context &  0.9587
          &   0.6589 
         &  0.7642
       & 3.2163
    &   0.4508
        & 0.5710
  &   0.1782
    \\


\ \ vanilla $MSE$ & 0.9580
          &   0.6567
         &  0.7537
       & 3.2441
    &   0,4544
        & 0.5722
  &   0.1795
    \\

\ \ no context &  0.9581
          &   0.6542  
         &  0.7595   
       & 3.1670
    &   0.4496 
        & 0.5645
  &   0.1764
    \\

\ \ \methodname & 0.9584
          &   0.6570
         &  0.7600 
       & 3.1879
    &   0.4472 
        & 0.5652
  &   0.1770
    \\


\midrule

\textit{Epistemic uncertainty} &  
          &     
         &     
       & 
    &    
        & 
  &   
    \\

\ \ vanilla $MSE$, no context &  0.9575
          &    0.6552
         &    0.7503
       & 3.2682
    &    0,4620
        & 0.5711
  &   0.1959
    \\

\ \ vanilla $MSE$, random context&  0.9524
          &    0.6459
         &    0.7427
       & 3.1028
    &    0.4422
        & 0.5498
  &   0.2068
    \\

\ \ vanilla $MSE$ &  0.9588
          &    0.6581
         &    0.7577
       & 3.2888
    &    \textbf{0.4661}
        & 0.5799
  &   0.1933
    \\

\ \ no context &  0.9592
          &    0.6630 
         &    \textbf{0.7744}
       & 3.2458
    &    0.4548
        & 0.5770
  &   0.1679
    \\


\ \ random context &  0.9599
          &    0.6577
         &    0.7588
       & 3.2479
    &    0.4566
        & 0.5739
  &  \textbf{0.1642}
    \\


\ \ \textbf{\methodname (ours) }&  \textbf{0.9605}
          &    \textbf{0.6654}
         &    $0.7723^{2nd}$
       & \textbf{3.3048}
    &    $0.4646^{2nd}$
        & \textbf{0.5843}
  &   $0.1690^{3rd}$
    \\
\bottomrule
\end{tabular}
\newline
\caption{Leave-one-subject-out ablation evaluation results using different evaluation metrics.
We present combinations of three ablation dimensions: uncertainty modelling, context modelling (either removing the context concatenation layer or by providing random context information), and vanilla mean squared error ($MSE$) instead of our proposed exponentially weighted $MSE$.}

\label{tab:results2}
\end{table*}

\begin{table*}[ht]

\centering
\begin{tabular}{llllllll}
\hline
Method & AUC-J $\uparrow$ & s-AUC $\uparrow$ & IG $\uparrow$ & NSS $\uparrow$ & SIM $\uparrow$ & CC $\uparrow$  & KLDiv $\downarrow$\\ \hline

MD-SEM~\cite{fosco2020much}  &    0.864
       &      \textbf{0.746}
        &      0.660
     &   \textbf{2.058}
  &      0.868
      &  0.774
  &   0.568
    \\

EMLNet~\cite{jia2020eml}  &    0.866
       &      0.746
        &      0.736
     &   2.050
  &      0.886
      &  0.780
  &   0.520
    \\
SAM-Res~\cite{cornia2018predicting}  &    0.865
       &      0.741
        &      0.538
     &   1.990
  &      0.899
      &  0.793
  &   0.610
    \\
ACNet-V17~\cite{li2021attention}  &    0.866
       &      0.739
        &      \textbf{0.854}
     &   1.948
  &      0.896
      &  0.786
  &   \textbf{0.228}
    \\
DI-Net~\cite{yang2019dilated}  &    0.862
       &      0.739
        &      0.195
     &   1.959
  &      \textbf{0.902}
      &  0.795
  &   0.864
    \\
MSI-Net~\cite{kroner2020contextual}  &    0.865
       &      0.736
        &      0.793
     &   1.931
  &      0.889
      &  0.784
  &   0.307
    \\
    GazeGAN~\cite{che2019gaze}  &    0.864
       &      0.736
        &      0.720
     &   1.899
  &      0.879
      &  0.773
  &   0.376
    \\
    FBNet~\cite{ding2021fbnet}  &    0.843
       &      0.706
        &      0.343
     &   1.687
  &      0.785
      &  0.694
  &   0.708
    \\
        SalFBNet-Res18~\cite{ding2021fbnet}  &    0.867
       &      0.733
        &      0.805
     &   1.950
  &      0.888
      &  0.773
  &   0.303
    \\
        SalFBNet-Res18Fixed~\cite{ding2021fbnet}  &    \textbf{0.868}
       &      0.740
        &      0.839
     &   1.952
  &      0.892
      &  0.772
  &   0.236
    \\
    \hline
\textbf{Context\text{-}free\text{-}SalNet (ours)} &    0.862 
       &      0.730
        &      0.750
     &   1.833
  &      0.763
      &  \textbf{0.870}
  &   0.308
    \\
    
    \hline
\end{tabular}
\newline
\caption{Evaluation results comparing Context-free-SalNET (since no context factors are presented for SALICON benchmark, we removed it from the architecture) to current top methods on SALICON free-viewing saliency benchmark~\cite{Jiang_2015_CVPR}. The lower values for KLDiv indicate better performance. For CC metric the values should approach either $1$ (positive correlation) or $-1$ (negative correlation), where $0$ means no correlation. The higher values, the better the performance rule is applied to the remaining metrics.}
\label{tab:results_salicon}
\end{table*}

Using our novel dataset, we evaluate \methodname on the task of pedestrian attention prediction both against state-of-the-art saliency prediction approaches as well as against ablations.
We also evaluate a context-free version of \methodname against state-of-the-art approaches on SALICON~\cite{jiang2015salicon} in order to estimate its performance on an established saliency benchmark dataset.
\vspace{-0.5cm}
\paragraph{Metrics.} In line with prior works~\cite{borji2012state,wang2018revisiting,bylinskii2014saliency}, we adopt the following evaluation metrics: AUC-Judd (AUC-J), AUC-Borji (AUC-B), shuffled AUC (s-AUC), Similarity Metric (SIM), Linear Correlation Coefficient (CC), Normalized Scanpath Saliency (NSS) and Kullback-Leibler Divergence (KLDiv)~\cite{bylinskii2018different}. 
\vspace{-0.5cm}
\paragraph{Comparison to SOTA saliency models.}
Table~\ref{tab:results1} shows the evaluation results of \methodname against the latest publicly available state-of-the-art approaches on the MIT benchmark\cite{mit-tuebingen-saliency-benchmark}, namely UniSal~\cite{droste2020unified}. 
We include the evaluation results of the pre-trained model provided by DeepGaze IIE~\cite{linardos2021deepgaze} (ranked 1st on MIT Saliency Benchmark) and SalFBNet~\cite{ding2021salfbnet} as training code is not publicly available.
Note however, that these results are not comparable to the other methods as DeepGaze IIE, for instance, uses several backbone networks and is trained on different datasets as well as it utilises center bias information computed on the target dataset.

\methodname clearly outperformed both the center bias baseline as well as UniSal~\cite{droste2020unified} (ranked 2nd on MIT Benchmark\footnote{\label{note1}https://saliency.tuebingen.ai/results.html}) across 6 out of 7 metrics.
\methodname clearly improves over DeepGaze2E in 5 out of 7 metrics, while it is close in AUC-B.

\vspace{-0.5cm}

\paragraph{Ablation Study.}
The results of our ablation study are summarized in Table~\ref{tab:results2}.
To quantify the effect of context modelling we created two different ablated versions: \textit{random context} consists of the exact same architecture as \methodname, but receives random context information as input.
For the \textit{no context} condition on the other hand we removed the context network and the context concatenation layer, resulting in fewer network parameters.
Crucially, \methodname clearly improves over both ablation conditions.
It improves in 6 out of 7 metrics over the random context condition and in 5 out of 7 metrics over the no context condition.
We also evaluated the impact of our novel ew-MSE loss by comparing to vanilla MSE.
Here, \methodname improved in 6 out of 7 metrics over the variant with vanilla MSE.
Finally, we observe clear improvement for epistemic uncertainty modelling. 
The ablation of \methodname without uncertainty modelling (i.e. no dropout at test time) is inferior in all 7 metrics.

\vspace{-0.5cm}
\paragraph{Performance on SALICON.}


While general saliency prediction is not the focus of this paper, we evaluate a context-free version of \methodname on SALICON\footnote{http://salicon.net/challenge-2017/} to obtain an estimate on how our architecture performs on this task in relation to SOTA approaches (see Table~\ref{tab:results_salicon}).
More precisely, \textit{Context-free-SalNET} consists of our encoder-decoder architecture including ew-MSE loss and epistemic uncertainty modelling, but without the context network and context concatenation layer.
Context-free-SalNet shows results that are close to the state of the art, and even outperfroms the other methods by a significant margin in the CC metric.

\subsection{Qualitative Evaluation}
In Figure~\ref{fig:best_worse}, we randomly selected success (top two rows) and failure cases (bottom two rows) of \methodname. 
DeepGaze2E strongly relies on center bias priors, leading to an over-estimation of the extent of the attention focus.
UniSal shows more accurate predictions in comparison to the ground-truth. 
SalFBNet models show comparable to UniSal results, but includes more false positive predictions. 
\methodname is able to produce predictions close to the ground truth without relying heavily on center bias or producing false positive predictions.
Columns 7-9 show qualitative results for the ablation conditions, supporting the utility of each method contribution. 
Rows 3-4 show samples of low-performing attention predictions, which holds across baseline and ablation evaluations for all models. 
Confirming the statement, that visual attention prediction for pedestrians in street-crossing scenarios is indeed a challenge due to the immense state-space of this navigation task.  

\section{Discussion}

\subsection{Applications}
Our method can be applied in all areas where precise modelling of pedestrian behavior is desired.
This includes driving simulators that can be used to train humans, but also the generation of training data for autonomous driving, where modelling and predicting pedestrian behavior is a key challenge~\cite{poibrenski2020m2p3}.
Furthermore, it can be used for critical scenario generation as an extension to~\cite{Vozniak1} in order to better understand and make predictions about dangerous traffic situations. 
Accurately modelling the attention of pedestrians in such scenarios can help to improve the generation of plausible walking trajectories~\cite{antakli2021hail}. Finally, by introducing certain extensions, it can be even applied to solve real-world problems as in~\cite{liang2020simaug}.

\subsection{Limitations and Future Work}
While we showed clear improvements of our methods over previous approaches, a number of aspects need to be addressed in future work.
While we evaluated the impact of context factors on pedestrian attention, in the future our approach should also be extended to include additional person-specific factors that are relevant in traffic scenarios~\cite{shen2012top,hadnett2019effect,dommes2015red,borji2014defending}. 
Joint attention between the driver of a car and the pedestrian is crucial in traffic situations~\cite{rasouli2017agreeing}, hence, an explicit representation could empower attention prediction models.
Furthermore, it will be important to include different roles of traffic participants (e.g. driver, bicyclist) in our model.
Additional challenges arise from the geographical location, since traffic scenarios can significantly differ throughout the world. Authors in~\cite{hell2021pedestrian} summarized significant cultural behaviour differences in street-crossing tasks between German and Japanese people. 
Moreover, while virtual reality is an effective research tool to collect close-to-natural data, future work also needs to find ways to validate results obtained in VR in the real world. 
The impact of different pre-trained weights, ModelNet for instance, on final performance is an interesting research question.

\vspace{-0.25cm}
\section{Conclusion}
\label{sec:conclusion}
We introduced \methodname, a novel context driven visual attention generation approach for street-crossing pedestrian scenarios. In evaluations of a newly recorded VR dataset of street crossing tasks including several task context factors, \methodname outperformed a state-of-the-art saliency prediction model and ablation experiments demonstrated our methods' ability to effectively exploit task context factors.
The dataset, including driving simulation setups and recorded gaze behavior will be made publicly available.
Together with our novel method, this dataset will be an important building block for future research on pedestrian attention prediction.
%
%
\vspace{-0.25cm}

\section{Acknowledgements}
This work has been funded by the German Ministry for Research and Education (BMBF) in the project REACT (grant no. 01IW17003). P. M\"uller was funded by BMBF (grant no. 01IS20075).

\label{sec:acknowledgement}

{\small
\bibliographystyle{ieee_fullname}
\bibliography{egbib}
}

\end{document}